\documentclass[10pt,letterpaper]{article}

\usepackage{hyperref}
\usepackage{cogsci}

\cogscifinalcopy 

\usepackage{pslatex}
\usepackage{float} 
\usepackage{subcaption}
\usepackage{booktabs}
\usepackage{graphicx}
\usepackage[dvipsnames]{xcolor}
\usepackage[normalem]{ulem}
\usepackage{multirow}
\usepackage{multicol}
\usepackage{array}
\usepackage{hhline}
\usepackage[poster]{tcolorbox}
\usepackage{etoolbox}
\usepackage{enumitem}
\usepackage{amssymb}
\usepackage{amsmath}
\AtBeginEnvironment{tcolorbox}{\small}

\usepackage{xurl}

\newcommand{\eg}{e.g.,~}
\newcommand{\ie}{i.e.,~}

\newcommand{\st}[1]{\bgroup\markoverwith{\textcolor{#1}{\rule[0.5ex]{2pt}{2pt}}}\ULon}

\newcolumntype{P}[1]{>{\raggedright\arraybackslash}p{#1}}

\renewcommand{\cite}{\shortcite}

\makeatletter
\renewcommand\paragraph{\@startsection{paragraph}{4}{\z@}%
                                    {1ex \@plus1ex \@minus.2ex}%
                                    {-1em}%
                                    {\normalfont\normalsize\bfseries}}

\makeatother
\usepackage{apacite}

\let\cite\shortcite

\title{Universal linguistic inductive biases via meta-learning}

\author{{\large \bf R. Thomas McCoy,\textsuperscript{1} Erin Grant,\textsuperscript{2} Paul Smolensky,\textsuperscript{3,1} Thomas L. Griffiths,\textsuperscript{4} and Tal Linzen\textsuperscript{1}} \\
  \texttt{tom.mccoy@jhu.edu}, \texttt{eringrant@berkeley.edu}, \texttt{smolensky@jhu.edu}, \texttt{tomg@princeton.edu}, \texttt{tal.linzen@jhu.edu} \\
  \textsuperscript{1}Department of Cognitive Science, Johns Hopkins University \\
  \textsuperscript{2}Department of Electrical Engineering \& Computer Sciences, University of California, Berkeley \\
  \textsuperscript{3}Microsoft Research AI, Redmond, WA USA \\
  \textsuperscript{4}Departments of Psychology and Computer Science, Princeton University 
}

\begin{document}

\maketitle

\begin{abstract}
How do learners acquire languages from the limited data available to them? 
This process must involve some inductive biases---factors that affect how a learner generalizes---but it is unclear which inductive biases can explain observed patterns in language acquisition.
To facilitate computational modeling aimed at addressing this question, we introduce a framework for giving particular linguistic inductive biases to a neural network model; such a model can then be used to empirically explore the effects of those inductive biases.
This framework disentangles universal inductive biases, which are encoded in the initial values of a neural network's parameters, from non-universal factors, which the neural network must learn from data in a given language.
The initial state that encodes the inductive biases is found with meta-learning, a technique through which a model discovers how to acquire new languages more easily via exposure to many possible languages.
By controlling the properties of the languages that are used during meta-learning, we can control the inductive biases that meta-learning imparts.
We demonstrate this framework with a case study based on syllable structure.
First, we 
specify the inductive biases that we intend to give our model, and then we translate those inductive biases into a space of languages from which a model can meta-learn. Finally, using existing analysis techniques, we verify that our approach has imparted the linguistic inductive biases that it was intended to impart.

\textbf{Keywords:} 
meta-learning, inductive bias, language universals, 
syllable structure typology, neural networks
\end{abstract}

\section{Introduction}

Human learners can acquire any of the world's languages from finite data. 
The acquisition of a particular language involves two factors: 
data from that language, 
and the learner's \textbf{inductive biases}, which are the factors that determine how the learner will generalize beyond the particular utterances in the data \cite{mitchell1997machine}.
Many inductive biases are shared by all humans (\eg because of shared brain anatomy or shared communicative goals), so these biases exert universal pressures on language acquisition. 
A central task of linguistics is to determine which inductive biases affect language acquisition and how those biases interact with learning to yield a learner's linguistic knowledge.\footnote{Though the term \textit{inductive biases} often refers to cognitive biases, we use it to encompass all pressures that shape the language that a learner learns; see Figure~\ref{fig:ug} and the Background section.}

\newcommand{\captionsquish}{
    \renewcommand{\baselinestretch}{0.8}\fontsize{8}{10}\selectfont}

\begin{figure}[t]
    \vspace{-15pt}
    \captionsetup[subfigure]{labelformat=empty,justification=justified,singlelinecheck=false}
    \begin{subfigure}{\columnwidth}
    \caption{\captionsquish\textbf{Step 1:} Translate a desired set of inductive biases into a space of languages.}
    \centering
    \par\smallskip
    \includegraphics[width=0.85\textwidth]{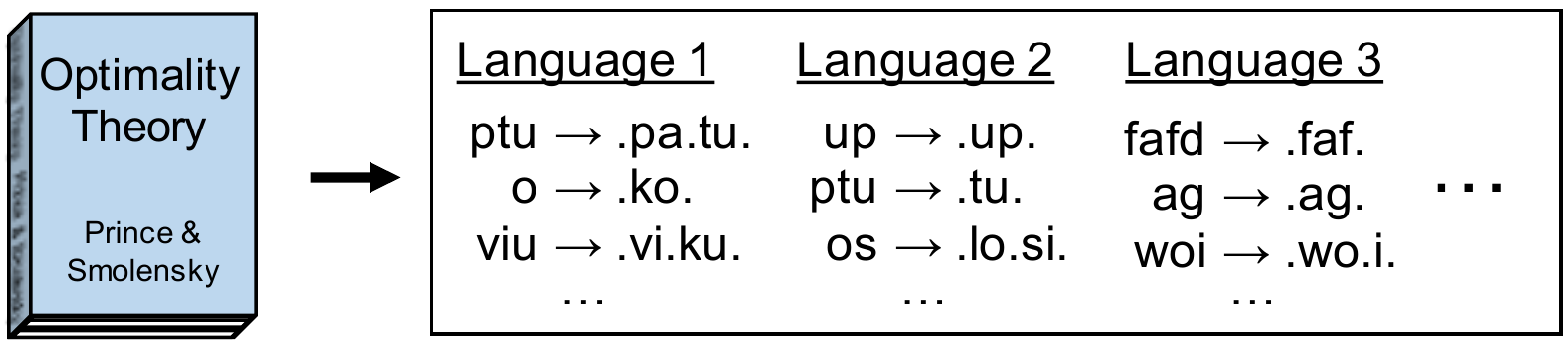}
    \label{fig:step1}
    \end{subfigure}
    \par\medskip
    \begin{subfigure}{\columnwidth}
    \caption{\captionsquish\textbf{Step 2:} Have a model ``meta-learn" from these languages to find a parameter initialization from which the model can acquire any language in the space.}
    \centering
    \par\smallskip
    \includegraphics[width=0.85\textwidth]{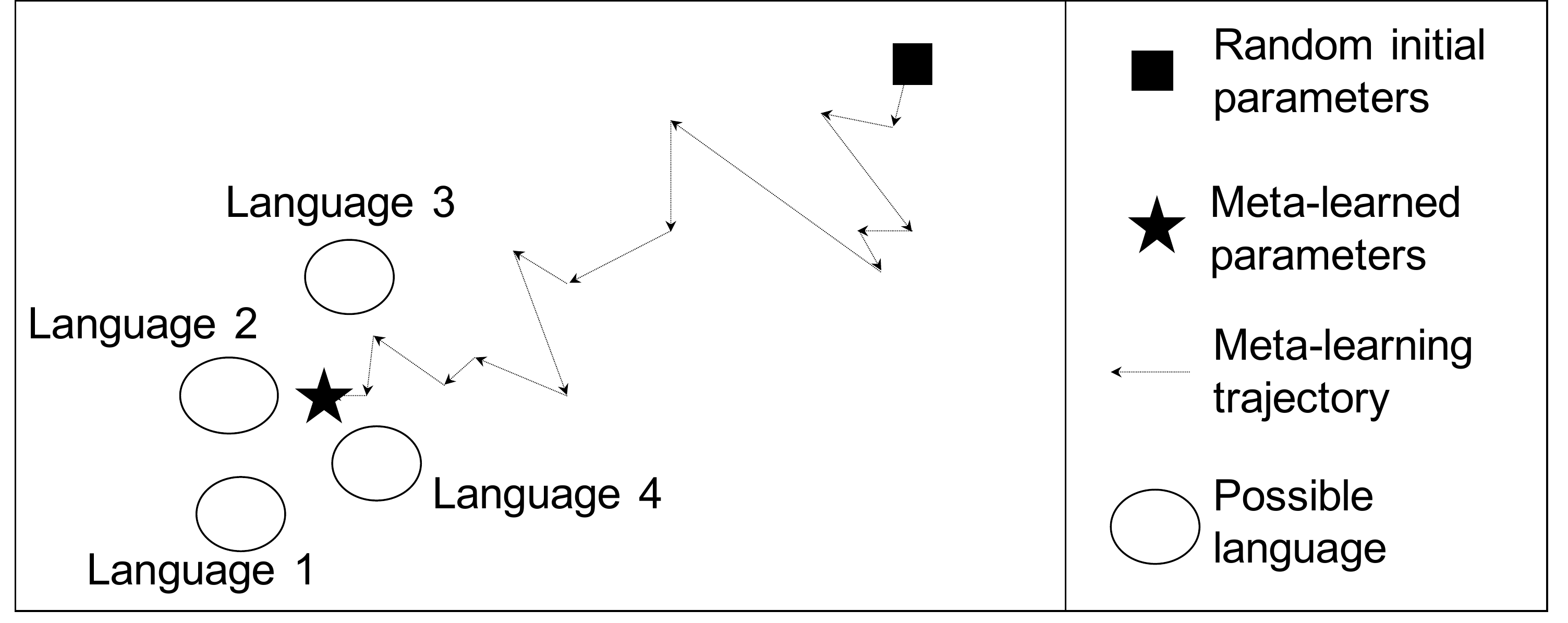}
    \label{fig:step2}
    \end{subfigure}
    \par\medskip
    \begin{subfigure}{\columnwidth}
    \caption{\captionsquish\textbf{Step 3:} Verify that meta-learning has imparted the desired inductive biases by training the model on analysis datasets.}
    \centering
    \par\smallskip
    \includegraphics[width=0.85\textwidth]{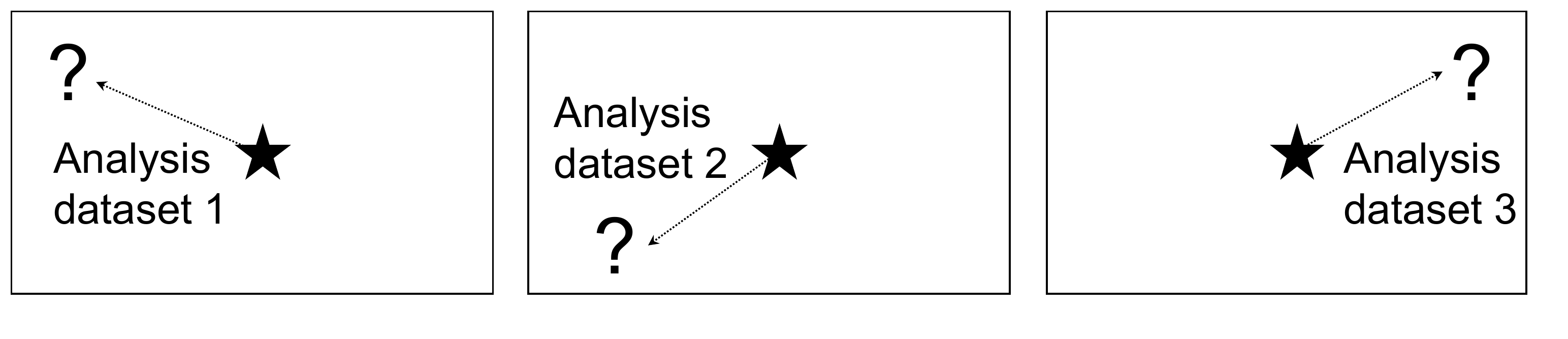}
    \label{fig:step3}
    \end{subfigure}
    \caption{Summary of our approach. The steps shown above give a pre-specified set of inductive biases to a model; alternately, Step~1 could be skipped by having the model meta-learn from existing languages, in which case the approach would discover a set of biases sufficient to acquire those existing languages rather than imparting a pre-specified set of biases (see Figure \ref{fig:metalearningnew}).}
    \label{fig:metalearning}
    \vspace{-12pt}
\end{figure}

There are two major approaches for modeling the interplay between universal 
inductive biases and learning. One approach, probabilistic modeling, typically follows a top-down methodology that commits to a  
representation and an
inference algorithm. Such strong commitments allow
targeted investigations (\eg \citeauthor{perfors2011learnability}, \citeyearNP{perfors2011learnability}) but can be too restrictive,
making it difficult for these models to represent all possible languages. In contrast, neural network modeling takes a bottom-up, data-driven approach that 
gives greater flexibility in representing the full span of 
languages. 
In recent work, neural networks trained on naturally-occurring data have shown 
success at learning
linguistic phenomena such as subject-verb agreement \cite{gulordava2018colorless}.
Since neural networks do not have overt biases specific to 
language, their successes give insight into which aspects of language are learnable from realistic input paired with domain-general  biases. 
However, these models often generalize in different ways from humans \cite{mccoy2019right}, and they require far more training data than humans, indicating that their learning is underconstrained  \cite{vanschijndel2019quantity}.
To address these problems, it would be necessary to give these models additional inductive biases that would appropriately constrain their learning to be more human-like, but their bottom-up nature makes it difficult to build in additional biases~\cite{griffiths2010probabilistic}.

In this work, we propose a computational modeling framework for imparting a hypothesized set of universal linguistic inductive biases
in a way that is compatible with the flexibility of neural networks.
Our approach is based on \textbf{meta-learning}, a
technique in which a learner is exposed to a variety of tasks, each of which comes with a limited amount of data~\cite{thrun1998learning,hochreiter2001learning}. This process instills in the learner a set of inductive biases which allow it to learn tasks similar to those it has seen before from limited data.
In our setting, each ``task" is a different language, and the inductive biases that result from meta-learning are encoded in a neural network's initial state.
This initial state is found in a data-driven manner; by controlling the data, we can influence which inductive biases will be encoded in the initial state, and the initial state can then be analyzed to verify that it encodes the universal inductive biases that it is intended to encode.

As a first case study, we show the effectiveness of this approach on the acquisition of a language's syllable structure, a paradigmatic example of universal linguistic inductive biases.
We define a set of inductive biases relating to syllable structure that we intend to give our model, and we then translate this set of inductive biases into a space of possible languages from which we have a model meta-learn. Through analysis of the meta-learned initial state, we verify that meta-learning has successfully imparted the inductive biases that it was intended to impart; for example, the model has meta-learned that the presence of certain input-output mappings in a language implies the presence of other input-output mappings.\footnote{Our code is at \url{https://github.com/tommccoy1/meta-learning-linguistic-biases}; there is also a demo at \url{http://rtmccoy.com/meta-learning-linguistic-biases.html}.}

\section{Background}

\definecolor{mygreen}{RGB}{178,252,178}
\definecolor{myblue}{RGB}{178,215,252}
\definecolor{mypurple}{RGB}{202,178,252}
\definecolor{mybluebox}{RGB}{4,8,115}

\tcbset{
  black/.style={colback=mybluebox,colframe=mybluebox!75!black},
  white/.style={colback=white,colframe=mybluebox!75!black},
  green/.style={colback=mygreen,colframe=mybluebox!75!black},
  blue/.style={colback=myblue,colframe=mybluebox!75!black},
  purple/.style={colback=mypurple,colframe=mybluebox!75!black},
}

\begin{figure}
  \vspace{-10pt}
  \begin{tcbposter}[
      poster = {
        columns = 11,
        rows = 8,
        height = .3\textheight,
        width = .48\textwidth,
        spacing=1mm,
      },
      boxes={
        size=small,
        fontupper=\scriptsize,
        valign=center,
        halign=center,
        segmentation style={solid, line width=1pt},
      },
    ]
    \posterbox[height=15pt,black]{yshift=-8.5pt, column=1, row=1, span=6}{\footnotesize\bfseries\textcolor{white}{Type of factor}}
    \posterbox[height=15pt,black]{yshift=-8.5pt, column=7, row=1, span=5}{\footnotesize\bfseries\textcolor{white}{Example}}

    \posterbox[green]{column=1, row=2, rowspan=5, span=2}{
        \textbf{Universal factors:} \\ Factors that are shared across all languages
      }
    \posterbox[green]{column=1, row=7, span=6, rowspan=2}{
      \textbf{Non-universal factors:} \\ Factors that vary across languages    
    }

    \posterbox[blue,boxsep=1pt]{column=3, row=2, rowspan=2, span=2}{Innate cognitive biases}
    \posterbox[blue]{column=3, row=4, span=4}{Physical and \\ perceptual constraints }
    \posterbox[blue]{column=3, row=5, span=4}{Functional pressures }
    \posterbox[blue]{column=3, row=6, span=4}{ Shared non-linguistic experience}

    \posterbox[purple,boxsep=1pt]{column=5, row=2, span=2}{Language-specific}
    \posterbox[purple]{column=5, row=3, span=2}{Domain-general}

    \posterbox[halign=flush left,white]{column=7, row=2, span=5}{Constraints on \textit{wh}-movement\\\cite{ross1967constraints}}
    \posterbox[halign=flush left,white]{column=7, row=3, span=5}{Simplicity bias\\\cite{perfors2011learnability}}
    \posterbox[halign=flush left,white]{column=7, row=4, span=5}{Vocal tract anatomy\\ \cite{maddieson1996phonetic}}
    \posterbox[halign=flush left,white]{column=7, row=5, span=5}{Communication efficiency\\ \cite{zipf1949human}}
    \posterbox[halign=flush left,white]{column=7, row=6, span=5}{Universality of some lexical concepts~\cite{swadesh1950salish}}
    \posterbox[halign=flush left,white]{column=7, row=7, span=5, rowspan=2}{
      Parameter settings in Principles and Parameters~\cite{chomsky1981lectures}; 
      
      \vspace{3pt}
      
      Constraint rankings in Optimality Theory~\cite{prince1993optimality} 
    }
  \end{tcbposter}%
\caption{Factors that shape languages and hypothesized examples.}
\label{fig:ug}
\vspace{-5pt}
\end{figure}

\begin{figure*}
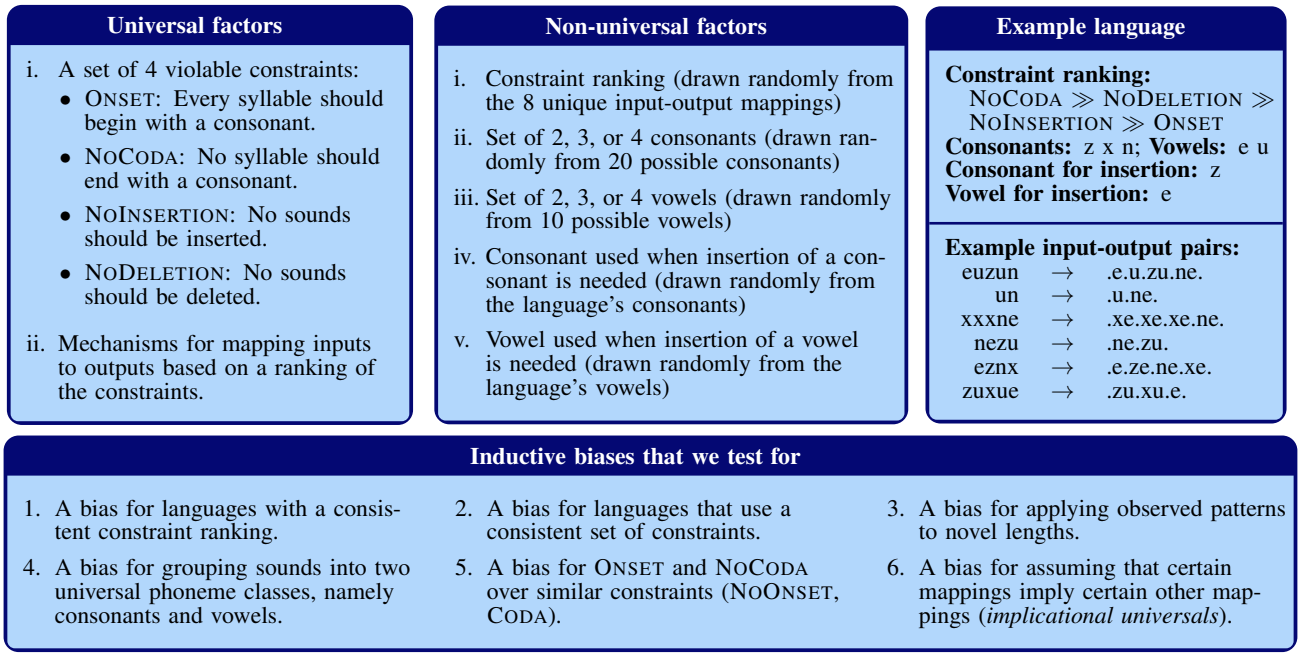

    \small
    \centering

    \begin{subfigure}{0.33\textwidth}
    \centering
    \begin{tcolorbox}[width=5.4cm, colback={myblue}, coltitle=white, toptitle=3pt, bottomtitle=3pt, colframe={mybluebox},outer arc=2mm,colupper=black, valign=center, halign=left, boxsep = 0pt, title={\textbf{Universal factors}}, halign title=center, segmentation style={solid, line width=1pt}, left=0pt]
        \def\arraystretch{1.0}%
    \begin{tabular}{p{0cm}p{4.5cm}}
         
        i. & A set of 4 violable constraints:
        \begin{itemize}[topsep=2pt]
            \setlength\itemsep{0pt}
            \item \textsc{Onset}: Every syllable should begin with a consonant.
            \item \textsc{NoCoda}: No syllable should end with a consonant.
            \item \textsc{NoInsertion}: No sounds should be inserted.
            \item \textsc{NoDeletion}: No sounds should be deleted.
        \end{itemize} \\
        ii. & Mechanisms for mapping inputs to outputs based on a ranking of the constraints. 
    \end{tabular}
    \end{tcolorbox}
    
    \end{subfigure}%
    \hfill
    \begin{subfigure}{0.33\textwidth}
    \centering
    \begin{tcolorbox}[width=6.3cm, colback={myblue}, coltitle=white, toptitle=3pt, bottomtitle=3pt, colframe={mybluebox},outer arc=2mm,colupper=black, valign=center, halign=left, boxsep = 0pt, title={\textbf{Non-universal factors}}, halign title=center, segmentation style={solid, line width=1pt}, left=0pt]
        \def\arraystretch{1.5}%
    \begin{tabular}{p{0cm}p{5.5cm}}
         
        i. & Constraint ranking (drawn randomly from the 8 unique input-output mappings)
         \\
        ii. & Set of 2, 3, or 4 consonants (drawn randomly from 20 possible consonants) \\ 
        iii. & Set of 2, 3, or 4 vowels (drawn randomly from 10 possible vowels) \\ 
        iv. & Consonant used when insertion of a consonant is needed (drawn randomly from the language's consonants) \\ 
        v. & Vowel used when insertion of a vowel is needed (drawn randomly from the language's vowels) \\ 
         
    \end{tabular}
        
        \end{tcolorbox}
    \end{subfigure}
    \hfill
    \begin{subfigure}{0.33\textwidth}
    \centering
    \begin{tcolorbox}[width=4.8cm, colback={myblue}, coltitle=white, toptitle=3pt, bottomtitle=3pt, colframe={mybluebox},outer arc=2mm,colupper=black, valign=center, halign=left, boxsep = 0pt, title={\textbf{Example language}}, halign title=center, segmentation style={solid, line width=1pt}, left=0pt]
        \begin{tabular}{p{4.5cm}}
        \hangindent=1em\textbf{Constraint ranking:} \newline\textsc{NoCoda} $\gg$ \textsc{NoDeletion} $\gg$ \textsc{NoInsertion}  $\gg$ \textsc{Onset} \\
        \textbf{Consonants:} z x n;
        \textbf{Vowels:} e u \\
        \textbf{Consonant for insertion:} z \\
        \textbf{Vowel for insertion:} e \\
        \end{tabular}
        
        \tcbline
        
        \small
        \def\arraystretch{1.0}%
       \begin{tabular}{rcl}
        \multicolumn{3}{l}{\textbf{Example input-output pairs:}} \\
         \footnotesize
        euzun & $\rightarrow$ & .e.u.zu.ne. \\
        un & $\rightarrow$ & .u.ne. \\
        xxxne & $\rightarrow$ & .xe.xe.xe.ne. \\
        nezu & $\rightarrow$ & .ne.zu. \\
        eznx & $\rightarrow$ & .e.ze.ne.xe. \\
        \color{myblue}X\color{black}zuxue & $\rightarrow$ & .zu.xu.e. \\
        \end{tabular}

        \end{tcolorbox}

    \end{subfigure}%
    
     \begin{subfigure}{\textwidth}
    \centering
    
    \begin{tcolorbox}[width=17.0cm, colback={myblue}, coltitle=white, toptitle=3pt, bottomtitle=3pt, colframe={mybluebox},outer arc=2mm,colupper=black, valign=center, halign=left, boxsep = 0pt, title={\textbf{Inductive biases that we  test for}}, halign title=center, segmentation style={solid, line width=1pt}, left=0pt, grow to left by=0.2cm,]
        \def\arraystretch{1.5}%
        \begin{tabular}{p{0cm}p{4.9cm}p{0cm}p{4.9cm}p{0cm}p{4.9cm}}
        1. & A bias for languages with a consistent constraint ranking. & 2. & A bias for languages that use a consistent set of constraints. & 3. & A bias for applying observed patterns to novel lengths. \\ 
        4. & A bias for grouping sounds into two universal phoneme classes, namely consonants and vowels. & 5. & A bias for \textsc{Onset} and \textsc{NoCoda} over similar constraints (\textsc{NoOnset}, \textsc{Coda}). & 6. & A bias for assuming that certain mappings imply certain other mappings (\textit{implicational universals}). \\ 
       \end{tabular}
    \end{tcolorbox}
    
    \end{subfigure}%
    
    \caption{A summary of basic syllable structure theory in Optimality Theory~\cite{prince1993optimality}. The top middle panel posits 8 unique input-output mappings because many of the 24 orderings of the 4 constraints are equivalent in the outputs they produce. The top right panel gives an example language; 
    the chosen constraint ranking leads to a language where no syllables end in a consonant, and where  violations of this restriction are fixed by vowel insertion rather than consonant deletion. Periods (present in the output but not the input) indicate syllable boundaries. The bottom panel lists the inductive biases that we use as behavioral tests of the universal factors.
    }
    \label{fig:otsummary}
    \vspace{-5pt}
\end{figure*}

\paragraph{Universal linguistic inductive biases}
Evidence for universal inductive biases that shape language acquisition primarily comes from two areas.  First, in typology (the taxonomy of observed language types), certain grammatical structures are much more common than others even across 
unrelated languages~\cite{greenberg1963universals}, and at least some of these patterns appear to arise from learners' inductive biases  \cite{culbertson2012learning}. Second, in acquisition, the argument from the poverty of the stimulus \cite{chomsky1980} notes that all language learners generalize in similar ways despite being faced with
stimuli that are consistent with multiple generalizations.
We use the phrase \textbf{universal linguistic inductive biases} for any pressures that universally affect acquisition, including the types of innate, language-specific constraints sometimes termed \textit{Universal Grammar}, 
as well as other influences such as articulatory or information-maximizing considerations; see Figure~\ref{fig:ug} for a categorization of universal pressures. We group these factors together because our
framework could be used to impart any type of inductive bias regardless of what source that bias might have in the real world.

Several linguistic formalisms provide theories of the universal/non-universal distinction. 
In the Principles and Parameters framework~\cite{chomsky1981lectures}, universal principles interact with non-universal parameter settings; 
in Optimality Theory~\cite{prince1993optimality}, a universal set of constraints interacts with non-universal rankings of these constraints. 
In contrast, our approach does not require any formal characterization of universal or non-universal factors; instead, due to the data-driven nature of the approach, these factors are characterized purely in terms of the behaviors they would lead to when paired with particular types of training data.
If a formal characterization of the model's inductive biases is desired, it must come from an analysis of the trained model, because the meta-learning process itself does not provide a formal characterization of the biases it imparts.

\paragraph{Learning and meta-learning}

The models we use are artificial neural networks, which are governed by a large number of numerical parameters such as connection weights.
At the core of our approach are two processes for determining the values of those parameters: \textbf{standard training} and \textbf{meta-training}. Standard training iteratively minimizes error within a single training language:
the model starts with a particular set of initial values for its parameters and is exposed to a training set  
of example input-output pairs from the language to be learned. For each training example, the output that the model generates is compared to the target value, and the model's parameters are adjusted to decrease
the difference between the predicted output and the correct target. Ideally, after many such updates, the model will perform well not only on its training set but also on a test set, which contains unseen examples drawn from the same language as the training set. 
Standard training requires the model to begin with some initial parameter values;
it is the task of meta-training to set these initial values based on data.

Standard training requires only a single language.
Meta-training, by contrast, samples multiple languages from a distribution of possible languages, $p(L)$. 
The particular form of meta-learning that we use is \emph{model-agnostic meta-learning}~(MAML; \citeNP{finn2017model}):
The model's initial state, $M_0$, is determined by a set of initial parameter values; then, for each sampled language $L_i \sim p(L)$, we train our model on the training set of $L_i$ using standard training,
to yield a trained model $M_i$. Crucially, we then compute an adjustment of the initial state $M_0$ using $M_i$'s loss on the unseen test examples from $L_i$; 
$M_i$ is discarded after the adjustment of $M_0$. 
Intuitively, we tweak $M_0$ so that, if we were to train the model on $L_i$ again, it would learn $L_i$ in fewer iterations. As meta-training proceeds, the initial model state $M_0$ (the square in Figure~\ref{fig:metalearning}, Step 2) moves to a point from which it can readily learn any language from the distribution of meta-training languages (the star in Figure~\ref{fig:metalearning}, Step 2).
We hypothesize that, if we 
construct $p(L)$ to encode the 
inductive biases that we wish our model to have, then meta-learning that aims to facilitate acquisition of languages in $p(L)$ will give a model these inductive biases.

\section{Overview of the approach}

The goal of our approach is to give a model a set of inductive biases hypothesized to be relevant for human cognition; once such a model has been created, it could then be used to empirically investigate the effects that those inductive biases have. 
The following sections walk through our 3-step approach, using a case study in syllable structure typology.

\subsection{Step 1: Defining the space of learning problems}

To apply our method, we must first define the inductive biases that we wish to impart; that is, we must define what innate knowledge we wish our model to have. For this purpose, we focus on the domain of syllable structure (the study of how words in a language are divided into syllables;~\citeauthor{jakobson1962selected}, \citeyearNP{jakobson1962selected}), and in particular we adopt the Optimality Theory account of \citeA{prince1993optimality},\footnote{We use the simplified account from Sec. 6.1 of \citeA{prince1993optimality}, leaving out the subsequent refinements.} because this account provides a clear characterization of which factors are universal and which are non-universal.

In this account, each word has an input form, which is the form it takes before any phonological processes have applied, and an output form, which is the result of phonological processes acting on the input.
For example, in English, the prefix \textit{in-} combines with 
the word \textit{possible} to create the input 
\textit{inpossible}, which is then mapped to the output 
\textit{impossible} 
through a place-assimilation process.
The input-output mapping
is determined by a ranked set of constraints, where the set of constraints is universal, but their ranking is non-universal. For syllable structure, we use four constraints. Two of them evaluate the output 
alone: \textsc{Onset} favors output syllables that begin with a consonant, and \textsc{NoCoda} favors  output syllables that end with a vowel.\footnote{A syllable's \textbf{onset} and \textbf{coda} consist of, respectively, syllable-initial and syllable-final consonants
(\eg for \textit{kep}, \textit{k} and \textit{p}, resp.).
} If the input does not satisfy these constraints (\eg \textit{kep}), the output could be made to satisfy them by either inserting or deleting phonemes (\eg \textit{.ke.pa.} or \textit{.ke.}).\footnote{We use periods (``.'') to indicate word and syllable boundaries.}
However, insertions and deletions are discouraged by the remaining constraints, \textsc{NoInsertion} and \textsc{NoDeletion}. 
When two constraints conflict with each other, the conflict is 
resolved by a priority-ranking of the constraints; this ranking differs across languages. In a language where \textsc{NoInsertion} and \textsc{NoDeletion} outrank \textsc{NoCoda}, the input \textit{kep} would map to the output \textit{.kep.}, because \textsc{NoCoda} cannot be satisfied
without violating a higher-ranked constraint. Under other rankings, the input \textit{kep} could map to \textit{.ke.}~or \textit{.ke.pa.}.

Based on this framework, we defined a set of inductive biases that we intend to give to our model via meta-learning (Figure \ref{fig:otsummary}, bottom panel). These biases were chosen to provide behaviorally-defined versions of the universal factors in the Optimality Theory framework. For example, this framework includes a universal mechanism for mapping inputs to outputs based on a constraint ranking, a mechanism which could not be directly observed in our model's behavior. Thus, we instead defined inductive biases that encode properties of this mechanism, such as a bias for languages with a consistent constraint ranking, to encode the fact that the  mechanism employs a consistent  ranking within each language.\footnote{Not all of these biases are necessarily present in humans; \eg languages differ in which phonemes can be syllabic nuclei, whereas one of our target biases is knowledge of a universal class of potential nuclei (\ie the vowels). We do not intend to propose a theory of syllable structure acquisition but rather to demonstrate how meta-learning could instantiate such a theory in a neural network.} 
We then translated these inductive biases into a space of possible languages and used that space to sample languages for use in meta-learning (Figure \ref{fig:otsummary}, top middle and top right panels).

\subsection{Step 2: Meta-training} \label{sec:phon_setup}

The next step is to train a meta-learner on the set of learning problems defined in Step 1. In our case study, the initial state $M_0$ and the language-specific state $M_i$ are the parameters of a sequence-to-sequence neural network~\cite{sutskever2014sequence} 
that maps a word's input form to a predicted output form. 
This architecture has two components: the \textbf{encoder} is fed the input one phoneme at a time and outputs a vector that encodes the entire input; this vector encoding is fed to the \textbf{decoder}, which generates the output 
symbols one at a time, ending
with
a special end-of-sequence token.\footnote{The particular model that we used was an LSTM \cite{hochreiter1997} with a single hidden layer of size 256, an embedding layer of dimension 10 (which learned distributed representations of phonemes), and no attention. The inner loop optimization of MAML used stochastic gradient descent with a learning rate of $1.0$ and batch size 100, while the outer loop optimization used Adam with a learning rate of $0.001$ \cite{kingma2015adam}. 
}
We apply meta-learning to such a model, allowing it to meta-learn from a set of 20,000 unique languages (called the \textit{meta-training set}). For each language, the model was trained on 100 examples from that language and then tested on 100 held-out examples; the model's meta-training objective
is thus
learning to perform \textit{100-shot learning}, that is, acquiring the ability to learn a new language from only 100 examples. After every set of 100 meta-training languages, we evaluated how well the model could perform 100-shot learning on each of 500 held-out languages;
we terminated meta-training when there had been 10 consecutive evaluations 
without improvement, and then evaluated the meta-trained model on its ability to perform 100-shot learning on a final set of 1,000 held-out languages
called the \textit{meta-test set}. 
Performance on 100-shot learning was measured as the proportion of inputs in a language's test set for which the model generated an exactly correct output sequence of phonemes and syllable boundaries after observing 100 examples from the language's training set.
We compared 
the model whose parameters were initialized using meta-learning to a baseline model whose parameters were initialized randomly; aside from 
initialization, both models use the same learning procedure to learn each language.

\paragraph{Meta-learning results} 
The model with meta-learned initial parameters had an average 100-shot accuracy
(\ie the accuracy after exposure to 100 examples) 
of 98.8\% on the languages in the meta-test set. 
By contrast, the 100-shot accuracy for a randomly-initialized model was only 6.5\%.
In this case study of syllable structure typology, then, meta-learning succeeded at imparting the ability to learn 
languages in our distribution of languages from a small number of examples. To evaluate whether meta-learning imparted the specific inductive biases that we intended it to impart (Figure \ref{fig:otsummary}), we next analyzed the weight initialization found through meta-learning by examining the learning behavior it produces.

\subsection{Step 3: Verification of the acquired inductive bias}

\paragraph{Ease of learning}
Our first approach for studying our model's inductive biases
is to evaluate how easily they learn 
languages that differ from each other in controlled ways.
We quantify ease of learning as the minimum number of training examples that a model needs from a 
language to reach 95\% accuracy on that language's test set.\footnote{Specifically, we selected the number of examples to be the smallest multiple of 100 for which the model converged to at least 95\% accuracy, without restricting the number of training iterations.}

We first use this technique to test whether the meta-learned inductive bias produces learning behavior that favors the \textbf{set of constraints} defining our syllable structure typology. 
Recall that our space of languages was defined with the constraints \textsc{Onset} and \textsc{NoCoda}. We now test our model on languages defined by these constraints, as well as languages defined by alternate constraint sets in which \textsc{Onset} is replaced with \textsc{NoOnset}, or \textsc{NoCoda} with \textsc{Coda}, or both.

\begin{figure}[t]
    \begin{subfigure}{\columnwidth}
    \centering
    \includegraphics[width=0.9\textwidth]{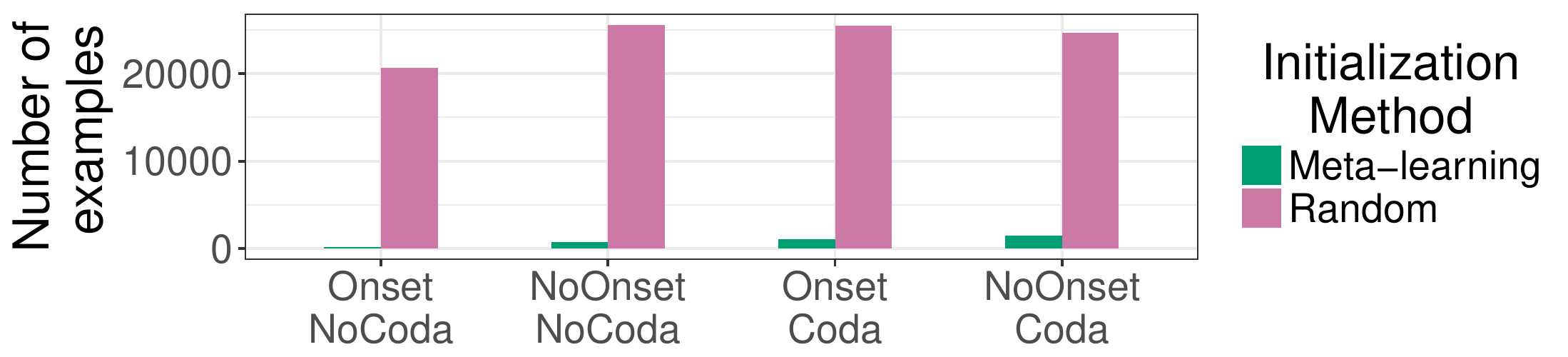}
    \label{fig:my_label}
    \end{subfigure}
    \begin{subfigure}{0.5\columnwidth}
    \centering
    \includegraphics[width=0.9\textwidth]{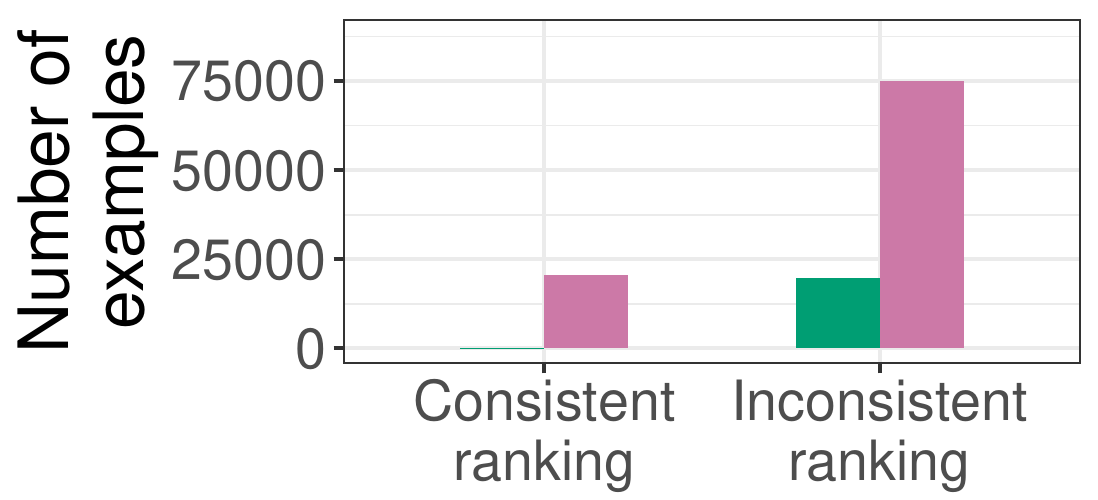}
    \label{fig:my_label}
    \end{subfigure}%
    \begin{subfigure}{0.5\columnwidth}
    \centering
    \includegraphics[width=0.9\textwidth]{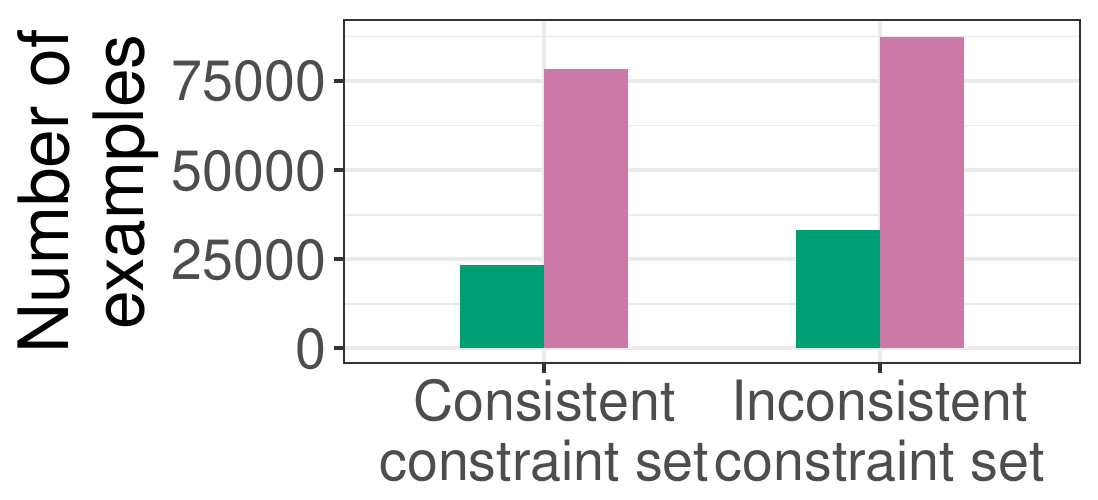}
    \label{fig:my_label}
    \end{subfigure}
    \caption{The number of examples needed to learn a language to 95\% accuracy (lower is better). Each bar is an average of 80 to 100 languages. Meta-learning improves performance on all conditions.
    }
    \label{fig:numexamples}
\end{figure}

\begin{figure}[t]
    \begin{subfigure}{0.32\columnwidth}
    \centering
    \includegraphics[width=\textwidth]{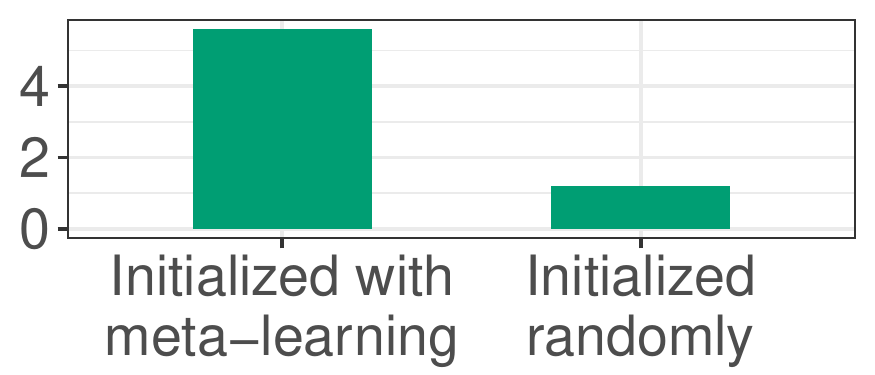}
    \caption{\captionsquish\raggedright Ratios comparing languages with an 
    \uline{incorrect set of constraints}
    to those with \textsc{Onset} and \textsc{NoCoda}.}
    \label{fig:ratioconstraints}
    \end{subfigure}
    \hfill
    \begin{subfigure}{0.31\columnwidth}
    \centering
    \includegraphics[width=\textwidth]{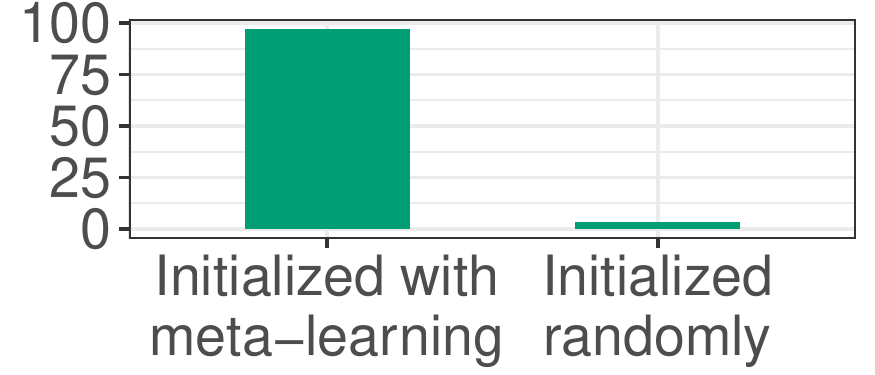}
    \caption{\captionsquish\raggedright Ratios comparing languages with an 
    \uline{inconsistent constraint ranking} to those with a consistent ranking.}
    \label{fig:ratioconsistentranking}
    \end{subfigure}%
    \hfill
    \begin{subfigure}{0.31\columnwidth}
    \centering
    \includegraphics[width=\textwidth]{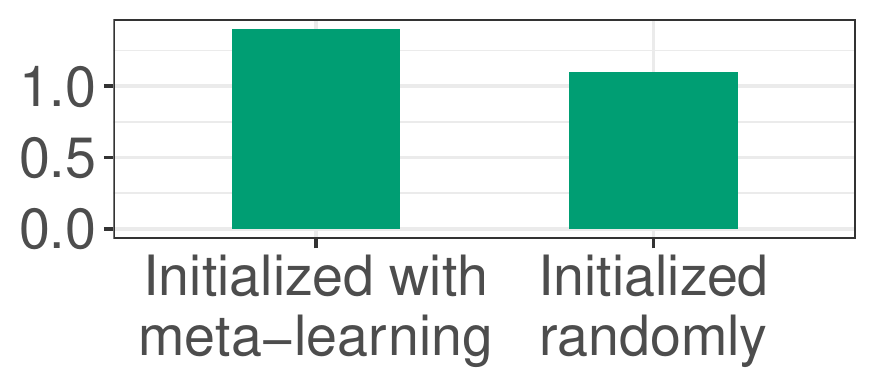}
    \caption{\captionsquish\raggedright Ratios comparing languages with an 
    \uline{inconsistent set of constraints} to those with a consistent set.}
    \label{fig:ratioconsistentset}
    \end{subfigure}
    \caption{
      Each subplot shows the ratio of the average number of examples need to learn a language with a property inconsistent with typology to the average number 
      needed to learn a language with the analogous consistent property;
      higher is better. In each case, the model initialized with meta-learning favors the typologically consistent language type more strongly than does the randomly initialized model. The ratios derive from the results shown in Figure~\ref{fig:numexamples}.}
    \label{fig:ratioexamples}
    \vspace{-10pt}
\end{figure}

Across language types, the model initialized with meta-learning required far fewer examples than the randomly-initialized model (Figure~\ref{fig:numexamples}, top). Importantly, though, meta-learning did not improve performance equally across languages:
The \textsc{Onset}/\textsc{NoCoda} languages were $5.6$ times easier to learn than languages defined by other constraints for the model initialized with meta-learning, compared to $1.2$ times easier for the random model
(Figure~\ref{fig:ratioconstraints}), suggesting
that meta-learning has imparted an inductive bias favoring languages that are consistent with the meta-training constraints.

\begin{figure}[t]
    \centering
    \includegraphics[width=0.9\columnwidth]{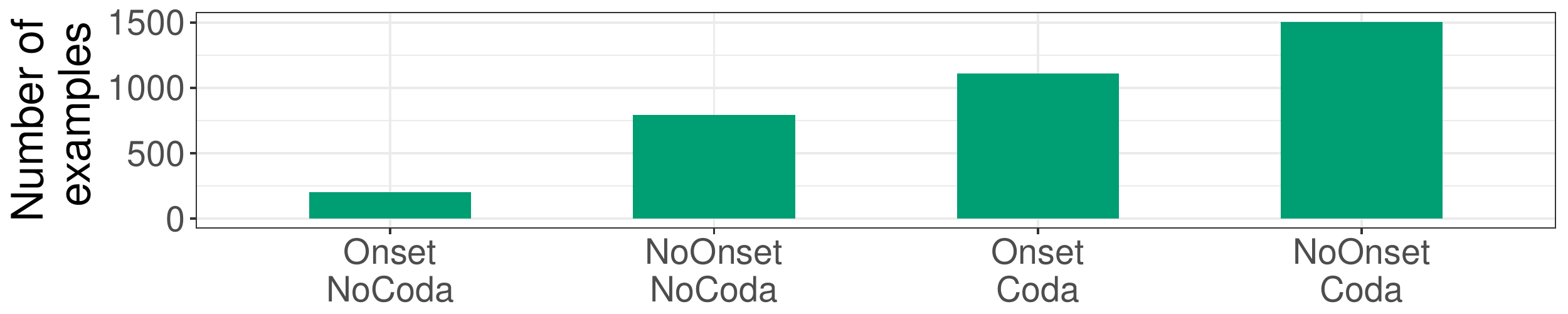}
    \caption{
        Data from the top panel of Figure~\ref{fig:numexamples} re-plotted
        at a different scale:
        The number of examples needed by the model initialized with meta-learning to learn languages with different sets of constraints.
      }
    \label{fig:mamlconstraintset}
    \vspace{-10pt}
\end{figure}

Has the model initialized with meta-learning simply memorized the types of languages it has seen, rather than learning the more abstract constraints of \textsc{Onset} and \textsc{Coda}? Figure~\ref{fig:mamlconstraintset} suggests that meta-learning has imparted some degree of more abstract knowledge, because, of the types of languages that were not present during meta-learning, the model has an easier time learning ones that have one of the correct constraints (\ie \textsc{NoOnset}/\textsc{NoCoda} languages and \textsc{Onset}/\textsc{Coda} languages) than ones that have neither correct constraint (\ie \textsc{NoOnset}/\textsc{Coda} languages).


We now test whether the model has meta-learned that \textbf{there must be a consistent constraint ranking within a language}. We test our models on languages governed by the constraints used during meta-learning (\textsc{Onset}, \textsc{NoCoda}, \textsc{NoInsertion}, and \textsc{NoDeletion}), but with no consistent ranking of constraints within the language. This is done by independently choosing a random constraint ranking for each input structure; \eg we might select a ranking for VV such that any input of the form VV maps to .CV.CV., while VVV might receive a ranking that maps VVV inputs to the empty string.\footnote{We use C and V as shorthands for \textit{consonant} and \textit{vowel}; CV is shorthand for \textit{any syllable of the form consonant-vowel}.}

For the model initialized with meta-learning, languages with a consistent ranking were 97.1 times easier to learn than languages without a consistent ranking, compared to only 3.6 times easier for the randomly initialized model
(Figure~\ref{fig:ratioconsistentranking}). This improvement in learning relative to random initialization suggests that meta-learning has strengthened the model's bias for languages generated by a consistent constraint ranking.

We next test whether our models have a bias for the fact that \textbf{within a language, a single set of constraints can consistently generate all input-output mappings} (the previous test was about the constraint \textit{ranking}, while this one is about the constraint \textit{set}). We  evaluate 
our models 
on languages with a consistent set of constraints but no consistent ranking across inputs, as in the previous experiment; but now we allow the set of constraints generating a given language to include any of the output constraint combinations discussed above (\textsc{Onset}/\textsc{Coda}, \textsc{Onset}/\textsc{NoCoda}, \textsc{NoOnset}/\textsc{Coda}, or \textsc{NoOnset}/\textsc{NoCoda}). We compare the learning of such languages to languages with no consistent set of constraints (and also no consistent constraint ranking), such that, for each input template (\eg CCVC), there is a randomly-selected set of constraints and a random ranking for those constraints.

On average, the languages with a consistent constraint set were 1.4 times easier to learn for the model initialized with meta-learning than languages without a consistent constraint set, compared to 1.1 times easier for the randomly initialized model (Figure~\ref{fig:ratioconsistentset}). This result suggests that meta-learning has moderately strengthened the model's bias favoring languages that can be generated by a single set of constraints. For such constraint-set consistency to greatly increase the learnability of a language, it appears necessary that the language also be generated by a single constraint ranking.

\begin{figure}
    \centering
    \includegraphics[width=\columnwidth]{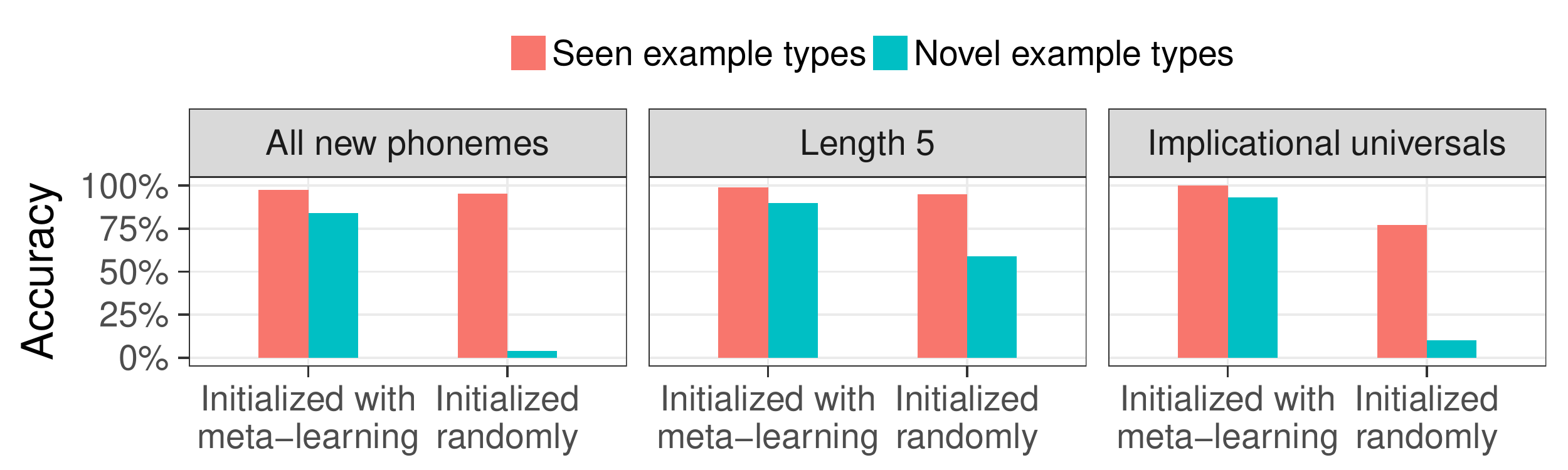}
    \caption{
    Results on poverty-of-the-stimulus experiments. Both models perform well on the example categories they have seen before, but the performance of the randomly-initialized model plummets when it is tested on novel types of examples; the meta-trained model exhibits less of a performance drop in these cases.
    }
    \label{fig:pos}
    \vspace{-10pt}
\end{figure}

\paragraph{Poverty of the stimulus}
As a second way to study
biases, we use a poverty-of-the-stimulus approach: for a given language, we train  
on a
dataset
lacking
a certain \textit{class} of examples, 
and test generalization to the withheld class. 

We performed three such experiments. In the \textbf{all new phonemes} setting, the training set for each language only contained 2 to 4 unique consonants and 2 to 4 unique vowels, as before, but now every example in the test set consisted entirely of consonants and vowels that were not present in the language's training set. The randomly initialized model 
has no hope of succeeding in this case, as it has no way to know whether each novel character is a consonant or vowel, but the model initialized with meta-learning could have learned these distinctions during meta-learning because the division between consonants and vowels is consistent across the meta-training languages. The model initialized with meta-learning performs strongly here (Figure~\ref{fig:pos}, left), suggesting that meta-learning has imparted universal consonant/vowel classes.

In the \textbf{length~5} setting, each language's training set only contained examples with an input length of at most 4, but its test set examples were all of length~5. The model initialized with meta-learning also performed strongly here (Figure~\ref{fig:pos}, middle). Note that, during meta-learning,
inputs with 
lengths up to
5 appeared; thus, the length 5 setting only requires generalization within the bounds seen during meta-learning. We also tested how models generalized from lengths at most 5 to length 6;
in this case, the model initialized with meta-learning only achieved 59\% accuracy. 
This suggests that meta-learning imparts a bias favoring languages in which the types of mappings that apply to short strings also apply to longer strings, but that this bias is restricted to the lengths present during meta-learning.

Last is the \textbf{implicational universals} setting. The space of languages that we have defined 
is restricted such that the presence of certain input-output mappings implies the presence of certain other 
input-output 
mappings.
For example, the presence of the mapping VC $\rightarrow$ .CVC. indicates that the language will insert a consonant at the start of any syllable that does not start with one, which means that 
the language will also have the mapping V $\rightarrow$ .CV.
To test whether a model has an inductive bias for this association, 
we can train it 
solely on examples of the form VC $\rightarrow$ .CVC. and then see how it handles V inputs; a naive model is unlikely to know how to handle this input, while a model that knows the implication would know to transform V to .CV. Our space of languages (Figure~\ref{fig:otsummary}) predicts 24 dependencies of this form; when we test  all of these dependencies, we find that the randomly-initialized model performs  poorly while the model initialized with meta-learning performs well (Figure~\ref{fig:pos}, right).  This suggests that the model has meta-learned these implicational universals.

\section{Conclusion}

We have demonstrated how meta-learning can impart universal inductive biases specified by the modeler.
This example-based approach to imparting inductive biases does not require an explicit theory of the biases in question; rather, imparting the biases only requires these biases to be translated into a distribution of possible languages. While the meta-learned biases are not as transparent as those encoded in probabilistic symbolic models,
analysis of the model's learning behavior can be used to evaluate whether 
meta-learning has produced
the desired biases, as we have shown. 
In our case study, we found evidence that meta-learning had successfully imparted all of our target inductive biases (or strengthened them, in cases where the biases were already present), including both some abstract biases (e.g., a bias for languages with a consistent constraint ranking) and some more concrete biases (e.g., a bias for treating certain phonemes as vowels).
These results show that linguistic inductive biases that have previously been framed in symbolic terms can be reformulated in the context of neural networks, facilitating cognitive modeling that combines the power of neural networks with the controlled inductive biases of symbolic approaches.

One important feature of the proposed approach is that it imparts soft biases rather than hard constraints. For example, after meta-learning, the model could learn attested language types more readily than unattested types---but it still could learn the unattested ones.
This capability is at odds with some theories that predict that unattested language types should be unlearnable, but there are reasons to believe that the consistent patterns seen in language typology and language acquisition may be best viewed as biases rather than constraints:
almost all linguistic universals have exceptions~(\citeauthor{evans2009myth}, \citeyearNP{evans2009myth}; though see \citeauthor{smolensky2009universals}, \citeyearNP{smolensky2009universals}); for example, the Arrernte language has been argued to be an exception to the syllable structure typology
we have adopted~\cite{breen1999arrernte}. Further, humans in artificial language learning experiments are capable of learning ``unnatural" languages \cite{moreton2012structure}.

\begin{figure}[t]
    \vspace{-15pt}
    \captionsetup[subfigure]{labelformat=empty,justification=justified,singlelinecheck=false}
    \begin{subfigure}{\columnwidth}
    \caption{\captionsquish\textbf{Step 1:} Have a model meta-learn from many natural languages.}
    \centering
    \par\smallskip
    \includegraphics[width=0.85\textwidth]{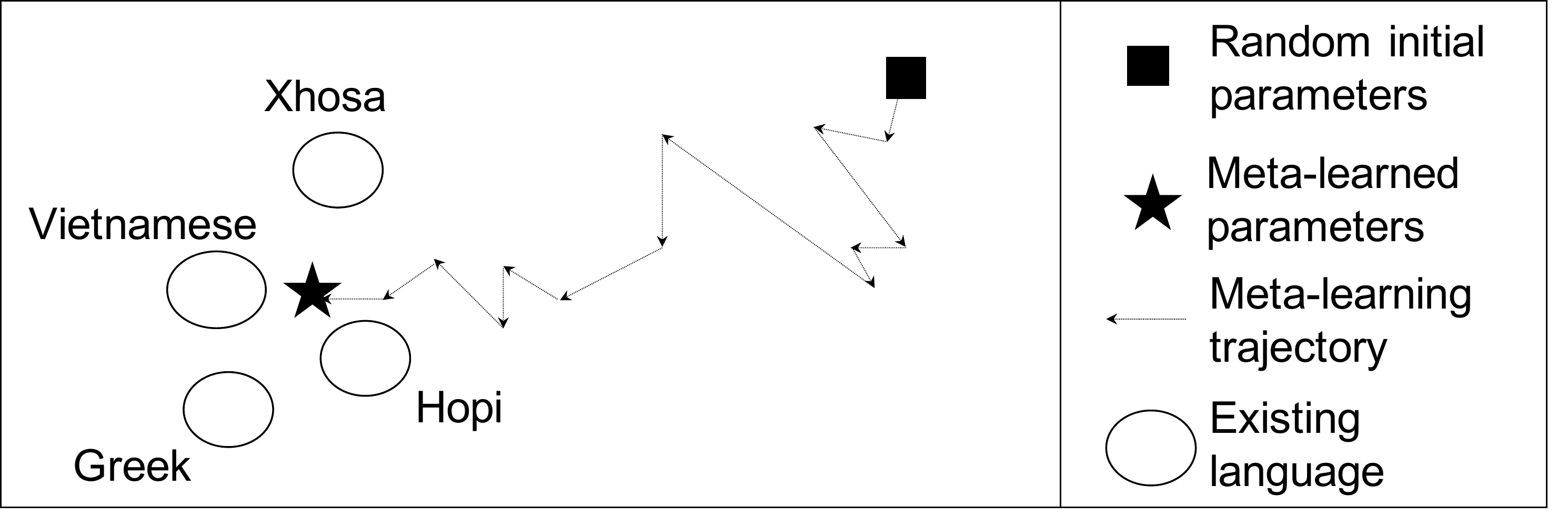}
    \label{fig:step2new}
    \end{subfigure}
    \par\smallskip
    \begin{subfigure}{\columnwidth}
    \caption{\captionsquish\textbf{Step 2:} Analyze the inductive biases that the model has meta-learned to gain insight into the biases that are relevant to the world's languages.}
    \centering
    \par\smallskip
    \includegraphics[width=0.85\textwidth]{metaug_step3b.pdf}
    \label{fig:step3new}
    \end{subfigure}
    \caption{Meta-learning from natural-language data.}
    \label{fig:metalearningnew}
    \vspace{-12pt}
\end{figure}

Several other works have discussed meta-learning from a cognitive perspective \cite{lake2017building,griffiths2019doing,lake2019compositional,grantlearning}, and in applied settings meta-learning has been 
applied to
language to create technology in low-resource languages~\cite{gu2018meta,ponti2019towards,kann2020learning}. Our novel contribution is the use of meta-learning to analyze the interplay between data and linguistic inductive biases.

Our approach can be used to test the behavioral effects of a particular inductive bias (\eg to test if the bias has the explanatory power hypothesized in a cognitive theory): All that is required to create a model with a specific inductive bias is a way to translate the bias into a distribution of meta-training languages, as we have demonstrated with Optimality Theory. 
In our experiments, we knew what factors defined the space of languages, and we showed that the inductive biases found through meta-learning reflected these factors; 
alternatively, this technique could be applied to naturally-occurring linguistic data for which we do not know the underlying data-generating process, to lend insight into the inductive biases that shaped this data (Figure \ref{fig:metalearningnew}). 
Finally, this framework is general enough that it can be straightforwardly applied to cognitive domains other than language (\eg vision).

\section{Acknowledgments}

For helpful comments, we are grateful to Colin Wilson, Paul Soulos, the members of the Johns Hopkins Computation and Psycholinguistics Lab, and the members of the Johns Hopkins Neurosymbolic Computation Lab. 
This research was supported by NSF Graduate Research Fellowship No. 1746891, NSF INSPIRE grant BCS-1344269, NSF grant BCS-1920924, and contract number FA8650-18-2-7832 from the Defence Advanced Research Projects Agency. Our experiments were conducted using the Maryland Advanced Research Computing Center (MARCC).

\bibliographystyle{apacite}

\bgroup
\setlength{\bibleftmargin}{.125in}
\setlength{\bibindent}{-\bibleftmargin}
\setlength{\itemsep}{0pt}
\bibliography{metaug}

\begin{thebibliography}{}

\bibitem [\protect \citeauthoryear {%
Breen%
\ \BBA {} Pensalfini%
}{%
Breen%
\ \BBA {} Pensalfini%
}{%
{\protect \APACyear {1999}}%
}]{%
breen1999arrernte}
\APACinsertmetastar {%
breen1999arrernte}%
\begin{APACrefauthors}%
Breen, G.%
\BCBT {}\ \BBA {} Pensalfini, R.%
\end{APACrefauthors}%
\unskip\
\newblock
\APACrefYearMonthDay{1999}{}{}.
\newblock
{\BBOQ}\APACrefatitle {Arrernte: A language with no syllable onsets} {Arrernte:
  A language with no syllable onsets}.{\BBCQ}
\newblock
\APACjournalVolNumPages{Linguistic Inquiry}{30}{1}{}.
\PrintBackRefs{\CurrentBib}

\bibitem [\protect \citeauthoryear {%
Chomsky%
}{%
Chomsky%
}{%
{\protect \APACyear {1980}}%
}]{%
chomsky1980}
\APACinsertmetastar {%
chomsky1980}%
\begin{APACrefauthors}%
Chomsky, N.%
\end{APACrefauthors}%
\unskip\
\newblock
\APACrefYearMonthDay{1980}{}{}.
\newblock
{\BBOQ}\APACrefatitle {Rules and representations} {Rules and
  representations}.{\BBCQ}
\newblock
\APACjournalVolNumPages{{BBS}}{3}{1}{}.
\PrintBackRefs{\CurrentBib}

\bibitem [\protect \citeauthoryear {%
Chomsky%
}{%
Chomsky%
}{%
{\protect \APACyear {1981}}%
}]{%
chomsky1981lectures}
\APACinsertmetastar {%
chomsky1981lectures}%
\begin{APACrefauthors}%
Chomsky, N.%
\end{APACrefauthors}%
\unskip\
\newblock
\APACrefYear{1981}.
\newblock
\APACrefbtitle {Lectures on government and binding} {Lectures on government and
  binding}.
\newblock
\APACaddressPublisher{Dordrecht}{Foris}.
\PrintBackRefs{\CurrentBib}

\bibitem [\protect \citeauthoryear {%
Culbertson%
, Smolensky%
\BCBL {}\ \BBA {} Legendre%
}{%
Culbertson%
\ \protect \BOthers {.}}{%
{\protect \APACyear {2012}}%
}]{%
culbertson2012learning}
\APACinsertmetastar {%
culbertson2012learning}%
\begin{APACrefauthors}%
Culbertson, J.%
, Smolensky, P.%
\BCBL {}\ \BBA {} Legendre, G.%
\end{APACrefauthors}%
\unskip\
\newblock
\APACrefYearMonthDay{2012}{}{}.
\newblock
{\BBOQ}\APACrefatitle {Learning biases predict a word order universal}
  {Learning biases predict a word order universal}.{\BBCQ}
\newblock
\APACjournalVolNumPages{Cognition}{122}{3}{}.
\PrintBackRefs{\CurrentBib}

\bibitem [\protect \citeauthoryear {%
Evans%
\ \BBA {} Levinson%
}{%
Evans%
\ \BBA {} Levinson%
}{%
{\protect \APACyear {2009}}%
}]{%
evans2009myth}
\APACinsertmetastar {%
evans2009myth}%
\begin{APACrefauthors}%
Evans, N.%
\BCBT {}\ \BBA {} Levinson, S\BPBI C.%
\end{APACrefauthors}%
\unskip\
\newblock
\APACrefYearMonthDay{2009}{}{}.
\newblock
{\BBOQ}\APACrefatitle {The myth of language universals: Language diversity and
  its importance for cognitive science} {The myth of language universals:
  Language diversity and its importance for cognitive science}.{\BBCQ}
\newblock
\APACjournalVolNumPages{{BBS}}{32}{5}{}.
\PrintBackRefs{\CurrentBib}

\bibitem [\protect \citeauthoryear {%
Finn%
, Abbeel%
\BCBL {}\ \BBA {} Levine%
}{%
Finn%
\ \protect \BOthers {.}}{%
{\protect \APACyear {2017}}%
}]{%
finn2017model}
\APACinsertmetastar {%
finn2017model}%
\begin{APACrefauthors}%
Finn, C.%
, Abbeel, P.%
\BCBL {}\ \BBA {} Levine, S.%
\end{APACrefauthors}%
\unskip\
\newblock
\APACrefYearMonthDay{2017}{}{}.
\newblock
{\BBOQ}\APACrefatitle {Model-agnostic meta-learning for fast adaptation of deep
  networks} {Model-agnostic meta-learning for fast adaptation of deep
  networks}.{\BBCQ}
\newblock
\BIn{} \APACrefbtitle {Proc. {ICML}.} {Proc. {ICML}.}
\PrintBackRefs{\CurrentBib}

\bibitem [\protect \citeauthoryear {%
Grant%
, Peterson%
\BCBL {}\ \BBA {} Griffiths%
}{%
Grant%
\ \protect \BOthers {.}}{%
{\protect \APACyear {2019}}%
}]{%
grantlearning}
\APACinsertmetastar {%
grantlearning}%
\begin{APACrefauthors}%
Grant, E.%
, Peterson, J\BPBI C.%
\BCBL {}\ \BBA {} Griffiths, T\BPBI L.%
\end{APACrefauthors}%
\unskip\
\newblock
\APACrefYearMonthDay{2019}{}{}.
\newblock
{\BBOQ}\APACrefatitle {Learning deep taxonomic priors for concept learning from
  few positive examples} {Learning deep taxonomic priors for concept learning
  from few positive examples}.{\BBCQ}
\newblock
\BIn{} \APACrefbtitle {{Proceedings of the 41st Annual Conference of the
  Cognitive Science Society}.} {{Proceedings of the 41st Annual Conference of
  the Cognitive Science Society}.}
\PrintBackRefs{\CurrentBib}

\bibitem [\protect \citeauthoryear {%
Greenberg%
}{%
Greenberg%
}{%
{\protect \APACyear {1963}}%
}]{%
greenberg1963universals}
\APACinsertmetastar {%
greenberg1963universals}%
\begin{APACrefauthors}%
Greenberg, J\BPBI H.%
\end{APACrefauthors}%
\unskip\
\newblock
\APACrefYear{1963}.
\newblock
\APACrefbtitle {Universals of language} {Universals of language}.
\newblock
\APACaddressPublisher{Cambridge, MA}{MIT Press}.
\PrintBackRefs{\CurrentBib}

\bibitem [\protect \citeauthoryear {%
Griffiths%
\ \protect \BOthers {.}}{%
Griffiths%
\ \protect \BOthers {.}}{%
{\protect \APACyear {2019}}%
}]{%
griffiths2019doing}
\APACinsertmetastar {%
griffiths2019doing}%
\begin{APACrefauthors}%
Griffiths, T\BPBI L.%
, Callaway, F.%
, Chang, M\BPBI B.%
, Grant, E.%
, Krueger, P\BPBI M.%
\BCBL {}\ \BBA {} Lieder, F.%
\end{APACrefauthors}%
\unskip\
\newblock
\APACrefYearMonthDay{2019}{}{}.
\newblock
{\BBOQ}\APACrefatitle {Doing more with less: meta-reasoning and meta-learning
  in humans and machines} {Doing more with less: meta-reasoning and
  meta-learning in humans and machines}.{\BBCQ}
\newblock
\APACjournalVolNumPages{Current Opinion in Behavioral Sciences}{29}{}{24--30}.
\PrintBackRefs{\CurrentBib}

\bibitem [\protect \citeauthoryear {%
Griffiths%
, Chater%
, Kemp%
, Perfors%
\BCBL {}\ \BBA {} Tenenbaum%
}{%
Griffiths%
\ \protect \BOthers {.}}{%
{\protect \APACyear {2010}}%
}]{%
griffiths2010probabilistic}
\APACinsertmetastar {%
griffiths2010probabilistic}%
\begin{APACrefauthors}%
Griffiths, T\BPBI L.%
, Chater, N.%
, Kemp, C.%
, Perfors, A.%
\BCBL {}\ \BBA {} Tenenbaum, J\BPBI B.%
\end{APACrefauthors}%
\unskip\
\newblock
\APACrefYearMonthDay{2010}{}{}.
\newblock
{\BBOQ}\APACrefatitle {Probabilistic models of cognition: Exploring
  representations and inductive biases} {Probabilistic models of cognition:
  Exploring representations and inductive biases}.{\BBCQ}
\newblock
\APACjournalVolNumPages{{TiCS}}{14}{8}{}.
\PrintBackRefs{\CurrentBib}

\bibitem [\protect \citeauthoryear {%
Gu%
, Wang%
, Chen%
, Li%
\BCBL {}\ \BBA {} Cho%
}{%
Gu%
\ \protect \BOthers {.}}{%
{\protect \APACyear {2018}}%
}]{%
gu2018meta}
\APACinsertmetastar {%
gu2018meta}%
\begin{APACrefauthors}%
Gu, J.%
, Wang, Y.%
, Chen, Y.%
, Li, V\BPBI O\BPBI K.%
\BCBL {}\ \BBA {} Cho, K.%
\end{APACrefauthors}%
\unskip\
\newblock
\APACrefYearMonthDay{2018}{}{}.
\newblock
{\BBOQ}\APACrefatitle {Meta-Learning for Low-Resource Neural Machine
  Translation} {Meta-learning for low-resource neural machine
  translation}.{\BBCQ}
\newblock
\BIn{} \APACrefbtitle {Proc. {EMNLP}.} {Proc. {EMNLP}.}
\PrintBackRefs{\CurrentBib}

\bibitem [\protect \citeauthoryear {%
Gulordava%
, Bojanowski%
, Grave%
, Linzen%
\BCBL {}\ \BBA {} Baroni%
}{%
Gulordava%
\ \protect \BOthers {.}}{%
{\protect \APACyear {2018}}%
}]{%
gulordava2018colorless}
\APACinsertmetastar {%
gulordava2018colorless}%
\begin{APACrefauthors}%
Gulordava, K.%
, Bojanowski, P.%
, Grave, E.%
, Linzen, T.%
\BCBL {}\ \BBA {} Baroni, M.%
\end{APACrefauthors}%
\unskip\
\newblock
\APACrefYearMonthDay{2018}{}{}.
\newblock
{\BBOQ}\APACrefatitle {Colorless Green Recurrent Networks Dream Hierarchically}
  {Colorless green recurrent networks dream hierarchically}.{\BBCQ}
\newblock
\BIn{} \APACrefbtitle {Proc. {NAACL}.} {Proc. {NAACL}.}
\PrintBackRefs{\CurrentBib}

\bibitem [\protect \citeauthoryear {%
Hochreiter%
\ \BBA {} Schmidhuber%
}{%
Hochreiter%
\ \BBA {} Schmidhuber%
}{%
{\protect \APACyear {1997}}%
}]{%
hochreiter1997}
\APACinsertmetastar {%
hochreiter1997}%
\begin{APACrefauthors}%
Hochreiter, S.%
\BCBT {}\ \BBA {} Schmidhuber, J.%
\end{APACrefauthors}%
\unskip\
\newblock
\APACrefYearMonthDay{1997}{}{}.
\newblock
{\BBOQ}\APACrefatitle {Long short-term memory} {Long short-term memory}.{\BBCQ}
\newblock
\APACjournalVolNumPages{Neural Computation}{9}{8}{1735--1780}.
\PrintBackRefs{\CurrentBib}

\bibitem [\protect \citeauthoryear {%
Hochreiter%
, Younger%
\BCBL {}\ \BBA {} Conwell%
}{%
Hochreiter%
\ \protect \BOthers {.}}{%
{\protect \APACyear {2001}}%
}]{%
hochreiter2001learning}
\APACinsertmetastar {%
hochreiter2001learning}%
\begin{APACrefauthors}%
Hochreiter, S.%
, Younger, A.%
\BCBL {}\ \BBA {} Conwell, P.%
\end{APACrefauthors}%
\unskip\
\newblock
\APACrefYearMonthDay{2001}{}{}.
\newblock
{\BBOQ}\APACrefatitle {Learning to learn using gradient descent} {Learning to
  learn using gradient descent}.{\BBCQ}
\newblock
\APACjournalVolNumPages{Proc. {ICANN}}{}{}{87--94}.
\PrintBackRefs{\CurrentBib}

\bibitem [\protect \citeauthoryear {%
Jakobson%
}{%
Jakobson%
}{%
{\protect \APACyear {1962}}%
}]{%
jakobson1962selected}
\APACinsertmetastar {%
jakobson1962selected}%
\begin{APACrefauthors}%
Jakobson, R.%
\end{APACrefauthors}%
\unskip\
\newblock
\APACrefYear{1962}.
\newblock
\APACrefbtitle {Selected Writings: Volume 1: Phonological Studies} {Selected
  writings: Volume 1: Phonological studies}.
\newblock
\APACaddressPublisher{}{Mouton}.
\PrintBackRefs{\CurrentBib}

\bibitem [\protect \citeauthoryear {%
Kann%
, Bowman%
\BCBL {}\ \BBA {} Cho%
}{%
Kann%
\ \protect \BOthers {.}}{%
{\protect \APACyear {2020}}%
}]{%
kann2020learning}
\APACinsertmetastar {%
kann2020learning}%
\begin{APACrefauthors}%
Kann, K.%
, Bowman, S\BPBI R.%
\BCBL {}\ \BBA {} Cho, K.%
\end{APACrefauthors}%
\unskip\
\newblock
\APACrefYearMonthDay{2020}{}{}.
\newblock
{\BBOQ}\APACrefatitle {Learning to Learn Morphological Inflection for
  Resource-Poor Languages} {Learning to learn morphological inflection for
  resource-poor languages}.{\BBCQ}
\newblock
\APACjournalVolNumPages{Proceedings of the Thirty-Fourth AAAI Conference on
  Artificial Intelligence}{}{}{}.
\PrintBackRefs{\CurrentBib}

\bibitem [\protect \citeauthoryear {%
Kingma%
\ \BBA {} Ba%
}{%
Kingma%
\ \BBA {} Ba%
}{%
{\protect \APACyear {2015}}%
}]{%
kingma2015adam}
\APACinsertmetastar {%
kingma2015adam}%
\begin{APACrefauthors}%
Kingma, D.%
\BCBT {}\ \BBA {} Ba, J.%
\end{APACrefauthors}%
\unskip\
\newblock
\APACrefYearMonthDay{2015}{}{}.
\newblock
{\BBOQ}\APACrefatitle {Adam: A Method for Stochastic Optimization} {Adam: A
  method for stochastic optimization}.{\BBCQ}
\newblock
\BIn{} \APACrefbtitle {Proc. {ICLR}.} {Proc. {ICLR}.}
\PrintBackRefs{\CurrentBib}

\bibitem [\protect \citeauthoryear {%
Lake%
}{%
Lake%
}{%
{\protect \APACyear {2019}}%
}]{%
lake2019compositional}
\APACinsertmetastar {%
lake2019compositional}%
\begin{APACrefauthors}%
Lake, B\BPBI M.%
\end{APACrefauthors}%
\unskip\
\newblock
\APACrefYearMonthDay{2019}{}{}.
\newblock
{\BBOQ}\APACrefatitle {Compositional generalization through meta
  sequence-to-sequence learning} {Compositional generalization through meta
  sequence-to-sequence learning}.{\BBCQ}
\newblock
\BIn{} \APACrefbtitle {Proc. {NeurIPS}.} {Proc. {NeurIPS}.}
\PrintBackRefs{\CurrentBib}

\bibitem [\protect \citeauthoryear {%
Lake%
, Ullman%
, Tenenbaum%
\BCBL {}\ \BBA {} Gershman%
}{%
Lake%
\ \protect \BOthers {.}}{%
{\protect \APACyear {2017}}%
}]{%
lake2017building}
\APACinsertmetastar {%
lake2017building}%
\begin{APACrefauthors}%
Lake, B\BPBI M.%
, Ullman, T\BPBI D.%
, Tenenbaum, J\BPBI B.%
\BCBL {}\ \BBA {} Gershman, S\BPBI J.%
\end{APACrefauthors}%
\unskip\
\newblock
\APACrefYearMonthDay{2017}{}{}.
\newblock
{\BBOQ}\APACrefatitle {Building machines that learn and think like people}
  {Building machines that learn and think like people}.{\BBCQ}
\newblock
\APACjournalVolNumPages{BBS}{40}{}{}.
\PrintBackRefs{\CurrentBib}

\bibitem [\protect \citeauthoryear {%
Maddieson%
}{%
Maddieson%
}{%
{\protect \APACyear {1996}}%
}]{%
maddieson1996phonetic}
\APACinsertmetastar {%
maddieson1996phonetic}%
\begin{APACrefauthors}%
Maddieson, I.%
\end{APACrefauthors}%
\unskip\
\newblock
\APACrefYearMonthDay{1996}{}{}.
\newblock
{\BBOQ}\APACrefatitle {Phonetic universals} {Phonetic universals}.{\BBCQ}
\newblock
\APACjournalVolNumPages{UCLA Working Papers in Phonetics}{}{}{}.
\PrintBackRefs{\CurrentBib}

\bibitem [\protect \citeauthoryear {%
McCoy%
, Pavlick%
\BCBL {}\ \BBA {} Linzen%
}{%
McCoy%
\ \protect \BOthers {.}}{%
{\protect \APACyear {2019}}%
}]{%
mccoy2019right}
\APACinsertmetastar {%
mccoy2019right}%
\begin{APACrefauthors}%
McCoy, R\BPBI T.%
, Pavlick, E.%
\BCBL {}\ \BBA {} Linzen, T.%
\end{APACrefauthors}%
\unskip\
\newblock
\APACrefYearMonthDay{2019}{}{}.
\newblock
{\BBOQ}\APACrefatitle {Right for the Wrong Reasons: Diagnosing Syntactic
  Heuristics in Natural Language Inference} {Right for the wrong reasons:
  Diagnosing syntactic heuristics in natural language inference}.{\BBCQ}
\newblock
\BIn{} \APACrefbtitle {Proc. {ACL}.} {Proc. {ACL}.}
\PrintBackRefs{\CurrentBib}

\bibitem [\protect \citeauthoryear {%
Mitchell%
}{%
Mitchell%
}{%
{\protect \APACyear {1997}}%
}]{%
mitchell1997machine}
\APACinsertmetastar {%
mitchell1997machine}%
\begin{APACrefauthors}%
Mitchell, T\BPBI M.%
\end{APACrefauthors}%
\unskip\
\newblock
\APACrefYearMonthDay{1997}{}{}.
\newblock
{\BBOQ}\APACrefatitle {Machine learning. 1997} {Machine learning. 1997}.{\BBCQ}
\newblock
\APACjournalVolNumPages{Burr Ridge, IL: McGraw Hill}{45}{37}{}.
\PrintBackRefs{\CurrentBib}

\bibitem [\protect \citeauthoryear {%
Moreton%
\ \BBA {} Pater%
}{%
Moreton%
\ \BBA {} Pater%
}{%
{\protect \APACyear {2012}}%
}]{%
moreton2012structure}
\APACinsertmetastar {%
moreton2012structure}%
\begin{APACrefauthors}%
Moreton, E.%
\BCBT {}\ \BBA {} Pater, J.%
\end{APACrefauthors}%
\unskip\
\newblock
\APACrefYearMonthDay{2012}{}{}.
\newblock
{\BBOQ}\APACrefatitle {Structure and substance in artificial-phonology
  learning, {Part II}: Substance} {Structure and substance in
  artificial-phonology learning, {Part II}: Substance}.{\BBCQ}
\newblock
\APACjournalVolNumPages{Language and linguistics compass}{6}{11}{}.
\PrintBackRefs{\CurrentBib}

\bibitem [\protect \citeauthoryear {%
Perfors%
, Tenenbaum%
\BCBL {}\ \BBA {} Regier%
}{%
Perfors%
\ \protect \BOthers {.}}{%
{\protect \APACyear {2011}}%
}]{%
perfors2011learnability}
\APACinsertmetastar {%
perfors2011learnability}%
\begin{APACrefauthors}%
Perfors, A.%
, Tenenbaum, J\BPBI B.%
\BCBL {}\ \BBA {} Regier, T.%
\end{APACrefauthors}%
\unskip\
\newblock
\APACrefYearMonthDay{2011}{}{}.
\newblock
{\BBOQ}\APACrefatitle {The learnability of abstract syntactic principles} {The
  learnability of abstract syntactic principles}.{\BBCQ}
\newblock
\APACjournalVolNumPages{Cognition}{118}{3}{}.
\PrintBackRefs{\CurrentBib}

\bibitem [\protect \citeauthoryear {%
Ponti%
, Vuli{\'c}%
, Cotterell%
, Reichart%
\BCBL {}\ \BBA {} Korhonen%
}{%
Ponti%
\ \protect \BOthers {.}}{%
{\protect \APACyear {2019}}%
}]{%
ponti2019towards}
\APACinsertmetastar {%
ponti2019towards}%
\begin{APACrefauthors}%
Ponti, E\BPBI M.%
, Vuli{\'c}, I.%
, Cotterell, R.%
, Reichart, R.%
\BCBL {}\ \BBA {} Korhonen, A.%
\end{APACrefauthors}%
\unskip\
\newblock
\APACrefYearMonthDay{2019}{}{}.
\newblock
{\BBOQ}\APACrefatitle {Towards Zero-shot Language Modeling} {Towards zero-shot
  language modeling}.{\BBCQ}
\newblock
\BIn{} \APACrefbtitle {Proc. {EMNLP-IJCNLP}.} {Proc. {EMNLP-IJCNLP}.}
\PrintBackRefs{\CurrentBib}

\bibitem [\protect \citeauthoryear {%
Prince%
\ \BBA {} Smolensky%
}{%
Prince%
\ \BBA {} Smolensky%
}{%
{\protect \APACyear {1993/2004}}%
}]{%
prince1993optimality}
\APACinsertmetastar {%
prince1993optimality}%
\begin{APACrefauthors}%
Prince, A.%
\BCBT {}\ \BBA {} Smolensky, P.%
\end{APACrefauthors}%
\unskip\
\newblock
\APACrefYear{1993/2004}.
\newblock
\APACrefbtitle {Optimality Theory: Constraint interaction in generative
  grammar} {Optimality theory: Constraint interaction in generative grammar}.
\newblock
\APACaddressPublisher{}{Wiley}.
\PrintBackRefs{\CurrentBib}

\bibitem [\protect \citeauthoryear {%
Ross%
}{%
Ross%
}{%
{\protect \APACyear {1967}}%
}]{%
ross1967constraints}
\APACinsertmetastar {%
ross1967constraints}%
\begin{APACrefauthors}%
Ross, J\BPBI R.%
\end{APACrefauthors}%
\unskip\
\newblock
\APACrefYear{1967}.
\newblock
\APACrefbtitle {Constraints on variables in syntax} {Constraints on variables
  in syntax}.
\newblock
\BUPhD, MIT.
\PrintBackRefs{\CurrentBib}

\bibitem [\protect \citeauthoryear {%
Smolensky%
\ \BBA {} Dupoux%
}{%
Smolensky%
\ \BBA {} Dupoux%
}{%
{\protect \APACyear {2009}}%
}]{%
smolensky2009universals}
\APACinsertmetastar {%
smolensky2009universals}%
\begin{APACrefauthors}%
Smolensky, P.%
\BCBT {}\ \BBA {} Dupoux, E.%
\end{APACrefauthors}%
\unskip\
\newblock
\APACrefYearMonthDay{2009}{}{}.
\newblock
{\BBOQ}\APACrefatitle {Universals in cognitive theories of language}
  {Universals in cognitive theories of language}.{\BBCQ}
\newblock
\APACjournalVolNumPages{{BBS}}{32}{5}{}.
\PrintBackRefs{\CurrentBib}

\bibitem [\protect \citeauthoryear {%
Sutskever%
, Vinyals%
\BCBL {}\ \BBA {} Le%
}{%
Sutskever%
\ \protect \BOthers {.}}{%
{\protect \APACyear {2014}}%
}]{%
sutskever2014sequence}
\APACinsertmetastar {%
sutskever2014sequence}%
\begin{APACrefauthors}%
Sutskever, I.%
, Vinyals, O.%
\BCBL {}\ \BBA {} Le, Q\BPBI V.%
\end{APACrefauthors}%
\unskip\
\newblock
\APACrefYearMonthDay{2014}{}{}.
\newblock
{\BBOQ}\APACrefatitle {Sequence to sequence learning with neural networks}
  {Sequence to sequence learning with neural networks}.{\BBCQ}
\newblock
\BIn{} \APACrefbtitle {Proc. {NeurIPS}.} {Proc. {NeurIPS}.}
\PrintBackRefs{\CurrentBib}

\bibitem [\protect \citeauthoryear {%
Swadesh%
}{%
Swadesh%
}{%
{\protect \APACyear {1950}}%
}]{%
swadesh1950salish}
\APACinsertmetastar {%
swadesh1950salish}%
\begin{APACrefauthors}%
Swadesh, M.%
\end{APACrefauthors}%
\unskip\
\newblock
\APACrefYearMonthDay{1950}{}{}.
\newblock
{\BBOQ}\APACrefatitle {Salish internal relationships} {Salish internal
  relationships}.{\BBCQ}
\newblock
\APACjournalVolNumPages{International Journal of American
  Linguistics}{16}{4}{}.
\PrintBackRefs{\CurrentBib}

\bibitem [\protect \citeauthoryear {%
Thrun%
\ \BBA {} Pratt%
}{%
Thrun%
\ \BBA {} Pratt%
}{%
{\protect \APACyear {1998}}%
}]{%
thrun1998learning}
\APACinsertmetastar {%
thrun1998learning}%
\begin{APACrefauthors}%
Thrun, S.%
\BCBT {}\ \BBA {} Pratt, L.%
\end{APACrefauthors}%
\unskip\
\newblock
\APACrefYear{1998}.
\newblock
\APACrefbtitle {Learning to learn} {Learning to learn}.
\newblock
\APACaddressPublisher{}{Kluwer Academic Publishers}.
\PrintBackRefs{\CurrentBib}

\bibitem [\protect \citeauthoryear {%
van Schijndel%
, Mueller%
\BCBL {}\ \BBA {} Linzen%
}{%
van Schijndel%
\ \protect \BOthers {.}}{%
{\protect \APACyear {2019}}%
}]{%
vanschijndel2019quantity}
\APACinsertmetastar {%
vanschijndel2019quantity}%
\begin{APACrefauthors}%
van Schijndel, M.%
, Mueller, A.%
\BCBL {}\ \BBA {} Linzen, T.%
\end{APACrefauthors}%
\unskip\
\newblock
\APACrefYearMonthDay{2019}{}{}.
\newblock
{\BBOQ}\APACrefatitle {Quantity doesn't buy quality syntax with neural language
  models} {Quantity doesn't buy quality syntax with neural language
  models}.{\BBCQ}
\newblock
\BIn{} \APACrefbtitle {Proc. {EMNLP-IJCNLP}.} {Proc. {EMNLP-IJCNLP}.}
\PrintBackRefs{\CurrentBib}

\bibitem [\protect \citeauthoryear {%
Zipf%
}{%
Zipf%
}{%
{\protect \APACyear {1949}}%
}]{%
zipf1949human}
\APACinsertmetastar {%
zipf1949human}%
\begin{APACrefauthors}%
Zipf, G\BPBI K.%
\end{APACrefauthors}%
\unskip\
\newblock
\APACrefYearMonthDay{1949}{}{}.
\newblock
{\BBOQ}\APACrefatitle {Human behavior and the principle of least effort.}
  {Human behavior and the principle of least effort.}{\BBCQ}
\newblock

\PrintBackRefs{\CurrentBib}

\end{thebibliography}
\egroup

\end{document}